\documentclass[letterpaper, 10 pt, conference]{ieeeconf}  %

\IEEEoverridecommandlockouts                              %

\usepackage[table]{xcolor}
\usepackage{graphicx}
\usepackage{cite}
\usepackage{multirow}
\usepackage{booktabs}
\usepackage{amsmath}
\usepackage{amssymb}
\usepackage{microtype}

\usepackage[utf8]{inputenc}
\usepackage[T1]{fontenc}
\usepackage{hyperref}

\DeclareMathOperator*{\argmax}{arg\,max}

\author{Vignesh Ramanathan \qquad\qquad Michael Milford \qquad\qquad Tobias Fischer}

\title{\LARGE \bf
Prepare for Warp Speed:\\Sub-millisecond Visual Place Recognition Using Event Cameras
}

\usepackage{eso-pic}
\usepackage{url}
\usepackage{svg}
\usepackage[capitalise]{cleveref}
\crefname{paragraph}{Paragraph}{Paragraphs}

\usepackage{graphicx}
\usepackage{subfigure}
\usepackage{float}
\usepackage{multirow}
\usepackage{booktabs}
\usepackage{graphicx}

\begin{document}

\bstctlcite{bstctl:forced_etal,bstctl:nodash}

\maketitle
\thispagestyle{empty}
\pagestyle{empty}

\begin{abstract}

Visual Place Recognition (VPR) enables systems to identify previously visited locations within a map, a fundamental task for autonomous navigation. Prior works have developed VPR solutions using event cameras, which asynchronously measure per-pixel brightness changes with microsecond temporal resolution. However, these approaches rely on dense representations of the inherently sparse camera output and require tens to hundreds of milliseconds of event data to predict a place. Here, we break this paradigm with Flash, a lightweight VPR system that predicts places using sub-millisecond slices of event data. Our method is based on the observation that active pixel locations provide strong discriminative features for VPR. Flash encodes these active pixel locations using efficient binary frames and computes similarities via fast bitwise operations, which are then normalized based on the relative event activity in the query and reference frames. Flash improves Recall@1 for sub-millisecond VPR over existing baselines by $11.33\times$ on the indoor QCR-Event-Dataset and $5.92\times$ on the 8 km Brisbane-Event-VPR dataset. Moreover, our approach reduces the duration for which the robot must operate without awareness of its position, as evidenced by a localization latency metric we term Time to Correct Match (TCM). To the best of our knowledge, this is the first work to demonstrate sub-millisecond VPR using event cameras.

\end{abstract}

\section{Introduction}
\label{sec:introduction}

Visual place recognition (VPR) is a fundamental capability for autonomous systems, allowing them to localize and navigate by recognizing previously visited locations~\cite{10261441, ijcai2021p603, 10937370, zhang2021visual, moskalenko2025visual}. For high-speed robots, achieving low-latency place matching is crucial to reduce the time without a position estimate and improve safety in applications such as self-driving vehicles~\cite{gehrig2024low} and facilitating GPS-denied navigation~\cite{8741612}. The latency of a VPR system arises from both sensing and processing delays. To address processing latency, traditional camera-based methods have investigated lightweight architectures, bioinspired models, and specialized visual cues to increase VPR speed while preserving accuracy~\cite{9981978, 10610918, 8972582}. However, conventional cameras remain limited by bandwidth-latency trade-offs, constraining their usability in mobile robotic applications. 

Event cameras are neuromorphic sensors that asynchronously detect per-pixel logarithmic changes in scene brightness. Unlike traditional cameras, event cameras sense the visual scene with microsecond resolution in a bandwidth-efficient manner~\cite{9138762}, enabling sub-millisecond perception and localization, which are critical for high-speed robotic applications. While some vision tasks leverage these advantages~\cite{10678150, gehrig2024low, dampfhoffer2025graph}, event camera-based VPR methods typically accumulate events over tens to hundreds of milliseconds to construct dense representations, thus failing to exploit the microsecond temporal resolution of event cameras~\cite{9201344, 9635907, 10197344, 9760407, 10802384, hines2025compact}. In this work, we challenge this paradigm by demonstrating that the \emph{spatial distribution of events alone}, even over sub-millisecond windows, contains sufficient discriminative information for place recognition.

\begin{figure}
    \centering
    \includegraphics[trim=0 0 0 0,clip,width=\linewidth]{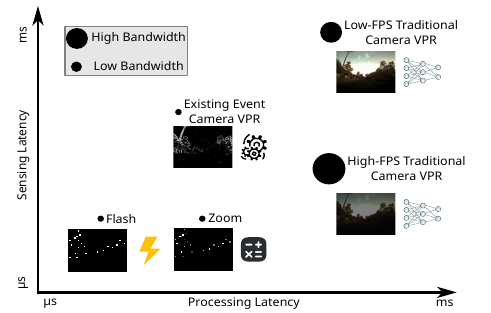}
    \vspace*{-0.5cm}
    \caption{Traditional cameras face a bandwidth–latency trade-off: higher frame rates demand more bandwidth, and processing dense frames is costly. Event cameras, by contrast, capture sparse scene changes at microsecond resolution. Existing methods, however, still accumulate tens to hundreds of milliseconds of events, squandering this speed. Flash exploits sub-millisecond slices of event data, using primarily bitwise operations to minimize both sensing and processing latencies.}
    \label{fig:fp}
    \vspace{-0.15cm}
\end{figure}

In this work, we introduce an efficient, ultra-low-latency VPR approach, termed \emph{Flash}. Our approach reduces sensing latency by relying on only sub-millisecond slices of event data. In the resulting event frames, fewer than 3\% of pixels are active per frame in the indoor dataset and fewer than 9\% in the outdoor dataset, with the average number of events per active pixel approaching one (\cref{subsubsec:ev_stat}). This makes a binary frame representation both a natural and a highly compact choice (\cref{fig:fp}).

While the binary representation exploits event sparsity, it also introduces an aliasing bias: reference frames with more active pixels tend to yield higher overlap scores, regardless of true similarity. To address this, we introduce a lightweight normalization that scales the similarity score by the ratio of active pixel counts between query and reference frames, mitigating the aliasing bias using only two floating-point operations per pair. Our design enables both sensing and processing to happen in a Flash.

Operating in the sub-millisecond regime poses a new challenge: the extremely short event windows produce a large number of reference frames, increasing both latency and computational cost. To address this, we propose uniformly sampling frames from the reference database to reduce its size while retaining sufficient map coverage. We further evaluate this strategy to characterize the trade-off between recognition accuracy and computational efficiency.

\noindent Our contributions can be summarized as follows:
\begin{enumerate}
    \item We introduce the first event camera-based VPR system that recognizes places using less than a millisecond of event data.
    \item We show that pixel locations responding to scene changes are sufficient for sub-millisecond place recognition with event cameras.
    \item We reduce the storage and search times of the large databases created by sub-millisecond accumulation windows through subsampling and analyze the resulting accuracy–compute trade-offs.
    \item We benchmark Flash against existing event camera-based VPR methods and conduct ablation studies to assess the sufficiency of active-pixel location encoding, the effectiveness of normalization in mitigating aliasing bias, and the distribution of time intervals during which the robot remains not localized.
\end{enumerate}

Flash’s sub-millisecond recognition capability enables new possibilities for time-critical localization, where traditional VPR methods fail to meet latency requirements. The code will be released publicly to support reproducibility.

\section{Related Works}
\label{sec:related_works}
This section reviews literature relevant to low-latency visual place recognition (VPR). We first discuss key works in traditional camera-based VPR that introduce concepts related to our approach, particularly visual saliency and efficient computing (\cref{subsec:traditional_camera_vpr}). We then review existing event camera-based VPR methods, showing how none have exploited the microsecond temporal resolution despite this being a key advantage of the sensor (\cref{subsec:event_camera_vpr}). Finally, we examine recent successes in sub-millisecond perception using event cameras in other vision tasks, demonstrating that ultra-low-latency processing is achievable but has not yet been realized for VPR (\cref{subsec:low_latency_app_event_camera}).

\subsection{Traditional Camera-based VPR}
\label{subsec:traditional_camera_vpr}

\subsubsection{Visual Saliency for VPR}
\label{subsubsec:visual_saliency_for_vpr}

Visual saliency, the perceptual quality that distinguishes certain scene regions, has been widely used in VPR to identify cues relevant for place recognition. Zaffar et al.~\cite{8972582} used entropy maps to select salient regions for Histogram-of-Gradients-based matching, while Wang et al.~\cite{9341703} performed saliency detection in the frequency domain for efficiency. Keetha et al.~\cite{9484750} reduced perceptual aliasing by using VLAD clusters to distinguish environment-specific from place-specific cues. Peng et al.~\cite{Peng_2021_ICCV} proposed a multi-level attention framework combining semantic guidance with attentional pyramid pooling, and Nie et al.~\cite{10528662} integrated saliency cues with semantic embeddings through pooling operations.

While effective, these methods require explicit saliency computation. In contrast, the ego-motion of an event camera naturally captures brightness changes, and the resulting event locations serve as strong discriminative features for place recognition.

\subsubsection{Efficient VPR}
\label{subsubsec:efficient_vpr}

Recent works have pursued computational efficiency to minimize VPR processing latency. Ferrarini et al.~\cite{9725251} binarized CNNs to a single-bit precision. They later addressed the full-precision input bottleneck with depthwise separable convolutions~\cite{9981978}. Grainge et al.~\cite{10494890} analyzed how network design choices impact recall-latency trade-offs.

Bio-inspired approaches have also shown promise: Arcanjo et al.~\cite{9672749,10439610} developed DorsoNet based on fruit fly circuits with voting mechanisms, while Hussaini et al.~\cite{9706280} and Hines et al.~\cite{10610918} employed SNNs with spike-timing-dependent plasticity and time-to-first-spike encoding to enable efficient place recognition.

Despite these advances, traditional camera-based methods remain fundamentally limited by frame capture rates. Event cameras overcome this limitation through microsecond-precision asynchronous sensing, enabling the ultra-low-latency recognition demonstrated in this work.

\subsection{Event Camera-Based VPR}
\label{subsec:event_camera_vpr}

Existing event camera-based place recognition methods have adapted events for use with NetVLAD~\cite{7937898} through various dense representations. Fischer et al.~\cite{9201344} binned events with multiple window sizes before reconstructing frames, Lee et al.~\cite{9635907} generated illumination-invariant edge images, and Lee et al.~\cite{10197344} used SNNs for edge reconstruction. Kong et al.~\cite{9760407} fused Event Spike Tensors~\cite{gehrig2019end} with NetVLAD. Although effective, these approaches process events as dense representations over tens- to hundreds-of-milliseconds windows, sacrificing the sparsity and microsecond temporal resolution inherent to event cameras. Processing the dense representations using NetVLAD introduces significant processing latencies, further bottlenecking the microsecond resolution of event cameras.

Other works have explicitly incorporated the unique sensing characteristics of event cameras into their designs. Fischer et al.~\cite{9760407} exploited the sparse event distribution by sampling pixel coordinates with high variability, improving computational efficiency. Hines et al.~\cite{hines2025compact} employed an SNN running on a SPECK\textsuperscript{TM} neuromorphic processor and demonstrates that combining neuromorphic sensing, algorithms, and hardware enables real-time, energy-efficient localization on a small, resource-constrained hexapod robot.

To date, no prior work has fully exploited the high temporal resolution of event cameras. Their change-driven, asynchronous events occur with microsecond precision, enabling the low-latency, bandwidth-efficient place recognition needed for high-speed robotics.

\subsection{Low-latency applications of event camera}
\label{subsec:low_latency_app_event_camera}

Prior works have leveraged the event camera's microsecond temporal resolution and spatiotemporally sparse output for low-latency perception. Gehrig et al.~\cite{gehrig2023recurrent} developed a recurrent vision architecture for object detection, achieving detection latencies below 10 ms on a commercial GPU. In a subsequent study, Gehrig et al.~\cite{gehrig2024low} fused 20 FPS traditional camera images with event camera data using a Graph Neural Network to achieve low latency without compromising detection accuracy in a bandwidth-efficient manner.

Many works have used hardware accelerators and neuromorphic chips to process event camera data to achieve sub-millisecond latency perception. Chiavazza et al.~\cite{Chiavazza_2023_CVPR} implemented a depth-from-motion algorithm on Intel's Loihi 2 neuromorphic chip, estimating optical flow directly from events via time-to-travel measurements and combining it with camera velocity to infer depth, achieving latencies below 0.5 ms. Dampfhoffer et al.~\cite{dampfhoffer2025graph} proposed a hybrid optical flow architecture with an asynchronous event branch for microsecond-scale predictions and a periodic branch for large-scale temporal context, breaking the accuracy-latency trade-off and achieving latencies of tens of microseconds on specialized hardware.

While other vision tasks have achieved sub-millisecond latencies with event cameras, VPR systems have yet to explore this capability. Flash fills this gap as the first event camera-based VPR system capable of recognizing places using less than a millisecond of event data, matching temporal capabilities demonstrated in other perception tasks.
\section{Preliminaries}
\label{subsec:preliminaries}

This section establishes the foundational concepts underlying our proposed method, Flash, providing the necessary background for understanding how it exploits event camera characteristics to achieve ultra-low-latency place recognition. We first formalize visual place recognition and its standard feature similarity matching formulation (\cref{subsec:vpr}). We then review event camera principles, focusing on their asynchronous and change-driven sensing model (\cref{subsec:event_cameras}).

\subsection{Visual Place Recognition (VPR)}
\label{subsec:vpr}

In Visual Place Recognition (VPR), the goal is to identify, for a given query frame $I_Q$, the reference frame $I_R \in \mathcal{D}$ that corresponds to the same physical location \cite{10261441}. Each frame is represented by a feature vector $\mathbf{f} \in \mathbb{R}^d$, extracted using a learned or handcrafted method.

The best-matching reference frame $I_R^*$ is determined by maximizing a similarity function $S(\cdot,\cdot)$:
\begin{equation}
I_R^* = \underset{I_R \in \mathcal{D}}{\argmax} \, S(\mathbf{f}_{I_Q}, \mathbf{f}_{I_R}).
\label{eq:vpr_similarity}
\end{equation}

Here, $\mathbf{f}_{I_Q}$ and $\mathbf{f}_{I_R}$ denote the feature vectors of the query 
and a reference frame, respectively. The function $S(\cdot,\cdot)$ typically 
represents a similarity measure such as cosine similarity.

\subsection{Event Cameras}
\label{subsec:event_cameras}

Event cameras asynchronously detect per-pixel brightness changes, resulting in bandwidth-efficient, 
high-temporal resolution perception of the environment \cite{4444573, 6889103}. Each detected change 
produces an event $e_k = (x_k, y_k, t_k, p_k)$, where $(x_k, y_k)$ are the spatial pixel coordinates, $t_k$ is the timestamp of the event, 
and $p_k \in \{+1, -1\}$ indicates the polarity of the intensity change (increase or decrease).

Given an event stream $\mathcal{E} = \{ e_k \}_{k=1}^N$,
we partition it into disjoint temporal windows $\mathcal{W}_i$ of duration $\delta$:
\begin{equation}
\mathcal{W}_i = \{ e_k \in \mathcal{E} \mid t_i < t_k \leq t_i+\delta \}.
\label{eq:temporal_window}
\end{equation}

From each window $\mathcal{W}_i$, an event frame $\mathbf{I}_i \in \mathbb{R}^{H \times W}$ can be 
constructed, where $(H, W)$ are the frame height and width. This binning process aggregates events into 
a frame-like representation (see \cref{eq:flash_binary_frame}).

\section{Methodology: Flash}
\label{sec:methodology}
We present Flash, a binary-frame-based visual place recognition system that achieves sub-millisecond latency by exploiting the sparsity of event data. Unlike existing methods that accumulate events over long time windows to form dense representations, Flash operates directly on the binary pattern of active pixels within very small window sizes.

\begin{figure}[!t]
    \centering
    \includegraphics[width=0.6\linewidth]{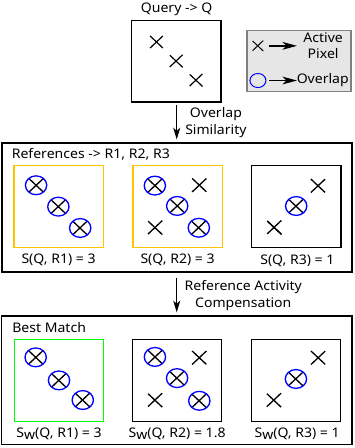}
    \vspace*{-0.3cm}
    \caption{\textbf{Overview of Flash:} We encode the active pixel locations as binary frames, and measure the query–reference similarity as their overlap using a bitwise operators. To prevent bias toward reference frames with high event activity, Reference Activity Compensation normalizes these similarity scores, filtering out incorrect matches.}
    \label{fig:flash}
    \vspace{-0.15cm}
\end{figure}

Flash, shown in \cref{fig:flash}, consists of three key components: (1) constructing binary frames that encode only pixel activity presence (\cref{subsec:flash_binary_frames}); (2) computing query-reference similarity through efficient bitwise operations on active pixels of query (\cref{subsec:flash_similarity}); and (3) applying Reference Activity Compensation (RAC) to prevent bias toward high-activity frames (\cref{subsec:rac}). This design enables Flash to perform reliable place recognition efficiently, using less than 1 ms of event data per query, with meaningful place recognition possible using as low as $\sim15\mu s$ of event data.

\subsection{Binary Frame Construction}
\label{subsec:flash_binary_frames}

Event streams are partitioned into disjoint temporal windows $\mathcal{W}_i$ of duration $\delta$ (see \cref{eq:temporal_window}). From each window, we construct a binary event frame 
$B_i \in \{0,1\}^{H \times W}$:
\begin{equation}
B_i(x,y) = \begin{cases}
1, & \text{if } \exists e_k \in \mathcal{W}_i \text{ such that } (x_k,y_k) = (x,y) \\
0, & \text{otherwise}.
\end{cases}
\label{eq:flash_binary_frame}
\end{equation}
This representation emphasizes the spatial distribution of the event activity while discarding intensity and frequency information, significantly reducing computational burden.

\subsection{Overlap Similarity}
\label{subsec:flash_similarity}

Let $\mathcal{B_Q}$ and $\mathcal{B_R}$ denote the binary query and reference frames, respectively. The set of active pixels in the query is defined as
\begin{equation}
\Omega_{\mathcal{B_Q}} = \{ (x,y) \mid \mathcal{B_Q}(x,y) = 1 \}.
\label{def:flash_active_query}
\end{equation}

The similarity between $\mathcal{B_Q}$ and $\mathcal{B_R}$ is defined as the number of overlapping active pixels, 
restricted to the query’s active set:
\begin{equation}
S(\mathcal{B_Q}, \mathcal{B_R}) = \sum_{(x,y) \in \Omega_{\mathcal{B_Q}}} \big( \mathcal{B_Q}(x,y) \,\&\, \mathcal{B_R}(x,y) \big),
\label{eq:flash_overlap_similarity}
\end{equation}
where $\&$ denotes the bitwise AND operator. This formulation ensures that only query-relevant locations contribute to the similarity score, with computational complexity $O(|\Omega_{\mathcal{B_Q}}|)$ rather than $O(H\cdot W)$, providing significant speedup when $|\Omega_{\mathcal{B_Q}}| \ll H\cdot W$. For the small time windows considered in Flash, $|\Omega_{\mathcal{B_Q}}|$ is typically in the tens, whereas $H\cdot W$ ranges from one to tens of thousands (see \cref{subsubsec:ev_stat}).

\subsection{Reference Activity Compensation (RAC)}
\label{subsec:rac}
Direct overlap similarity (\cref{eq:flash_overlap_similarity}) favors reference frames with high activity, which naturally produce more overlapping pixels regardless of actual scene correspondence (see \cref{fig:flash}). To address this bias, we introduce Reference Activity Compensation (RAC), a lightweight normalization scheme.

\begin{equation}
\tilde{w}(\mathcal{B_Q}, \mathcal{B_R}) = 
\begin{cases}
\frac{|\Omega_{\mathcal{B_Q}}|}{|\Omega_{\mathcal{B_R}}|}, & \text{if } |\Omega_{\mathcal{B_R}}| > |\Omega_{\mathcal{B_Q}}| \\
1, & \text{otherwise},
\end{cases}
\label{eq:rac}
\end{equation}

Here, $|\Omega_{\mathcal{B_R}}|$ denotes the number of active pixels in reference frame $\mathcal{R}$. Reference frames with large $|\Omega_{\mathcal{B_R}}|$ are more likely to alias with a query frame $\mathcal{B_Q}$, and the above normalization mitigates this effect when the query frame does not exhibit comparable event activity. When $|\Omega_{\mathcal{B_R}}| \leq |\Omega_{\mathcal{B_Q}}|$, no adjustment is applied, as RAC targets aliasing from overly active reference frames; applying the ratio in this case could bias the query toward sparser frames.

The normalized similarity is then computed as:
\begin{equation}
S_w(\mathcal{B_Q}, \mathcal{B_R}) = \tilde{w}(\mathcal{B_Q}, \mathcal{B_R}) \cdot S(\mathcal{B_Q}, \mathcal{B_R}),
\label{eq:flash_weighted_similarity}
\end{equation}
where $S(\mathcal{B_Q}, \mathcal{B_R})$ is the raw overlap similarity from \cref{eq:flash_overlap_similarity}. This adjustment penalizes overly active reference frames, ensuring a balanced comparison with only two floating-point operations per query-reference pair.

Finally, the best-matching reference frame is selected by maximizing the weighted similarity (following \cref{eq:vpr_similarity}):
\begin{equation}
B_R^* = \underset{\mathcal{B_R} \in \mathcal{D}}{\text{argmax}}\  S_w(\mathcal{B_Q}, \mathcal{B_R}).
\label{eq:flash_best_match}
\end{equation}

\subsection{System Efficiency}
\label{subsec:implementation}
By restricting computations to active query pixels and using lightweight operations, Flash delivers a fast and efficient solution for sub-millisecond event-based VPR. The binary-frame design ensures scalability, bitwise operations for similarity guarantee low processing times, and overlap normalization improves robustness under varying levels of scene activity.

Flash's design enables highly efficient implementation through several key optimizations. The sparse binary arrays of the reference database can be stored efficiently by keeping only the coordinates of the active pixels, which are few for small window sizes. The similarity computation can leverage hardware-accelerated bitwise AND operations and population count (popcount) instructions available on modern CPUs for fast and cheap computations. RAC normalization requires only two floating-point operations: one for the normalization factor and one for multiplying the overlap similarity. Together, these implementation strategies can massively reduce the processing latencies over existing approaches and allow the high sensing capabilities of event cameras to be fully utilized.

\section{Experimental setup}
\label{subsec:setup}

This section describes the datasets used to evaluate our approach (\cref{subsec:datasets}) and defines the performance metrics (\cref{subsec_eval_mets}). including a new metric, Time to Correct Match (TCM), which captures the temporal dynamics of place recognition. We benchmark our proposed approach against existing methods and introduce a strong baseline, termed \emph{Zoom}, to assess both performance and efficiency (\cref{subsubsec:baselines}).

\subsection{Datasets}
\label{subsec:datasets}

We evaluate on two event-based VPR datasets recorded using the DAVIS346 camera. The Brisbane-Event-VPR dataset (henceforth \emph{Brisbane}) spans an 8 km urban route traversed six times at $\sim$15 m/s, with natural stops and varying illumination and weather (excluding night-time). The QCR-Event-VPR dataset (henceforth \emph{QCR}) covers a 160 m indoor route traversed 16 times at $\sim$1 m/s with different speeds and camera orientations. Together, Brisbane and QCR enable evaluation across large-scale outdoor driving and structured indoor robotics.

\subsection{Evaluation Metrics}
\label{subsec_eval_mets}

\subsubsection{Recall@1}
\label{subsubsec:recall1}

To evaluate matching performance, we use Recall@1, which measures the fraction of query frames for which the top-ranked reference frame is a correct match. 
Formally, Recall@1 is defined as
\begin{equation}
\text{Recall@}1 = \frac{1}{N} \sum_{n=1}^{N} 
\mathbf{1}\big( I_R^* \in \mathcal{M}(I_Q^n) \big),
\label{eq:recall1}
\end{equation}
where $N$ is the number of query frames, $I_R^*$ denotes the top-ranked reference frame for query $I_Q^n$, 
$\mathcal{M}(I_Q^n)$ is the set of ground-truth matches for query $I_Q^n$, 
and $\mathbf{1}(\cdot)$ is the indicator function.

\subsubsection{Time between Correct Match (TCM)}
\label{subsubsec:tcm}

We introduce TCM as a metric to evaluate the temporal performance of a VPR system. In environments where the robot encounters no unseen places, TCM quantifies the duration between consecutive correct place recognitions during a traverse.

For a sequence of queries $\{I_Q^1, \dots, I_Q^N\}$, let 
\begin{equation}
\delta_n = 
\begin{cases} 
1, & \text{if the system correctly matches } I_Q^n \\
0, & \text{otherwise}
\end{cases}
\label{eq:delta}
\end{equation}
and define the time to the $i$-th correct match as
\begin{equation}
\text{TCM}_i = t_i - t_{i-1}, \quad t_i = \min \{ n > t_{i-1} \mid \delta_n = 1 \},  t_0 = 0.
\label{eq:tcm}
\end{equation}

Using the set of TCM values from a traverse, we can construct a TCM distribution that measures the number of correct matches for the given TCM time over the complete traverse:
\begin{equation}
P(\text{TCM} = \tau) = \frac{\# \text{ of correct matches with TCM } \tau}{\text{total number of places to match}}. 
\label{eq:tcm_dist}
\end{equation}

This distribution can be converted to a Cumulative Distribution Function (CDF) and it provides a measure of the time durations during which the robot is unable to obtain a correct position estimate. The higher the CDF value for a given TCM, the shorter the period of position ``blindness''. Unlike Recall@1, which measures average performance over the entire traverse, the CDF provides a more granular measure of VPR performance. This new measure is a more useful indicator of utility in latency-sensitive scenarios when the correct position is required within a specific time window, like fast moving robotic platforms, vehicles and drones.

\begin{figure*}[!ht]
  \centering
  \includegraphics[width=\linewidth]{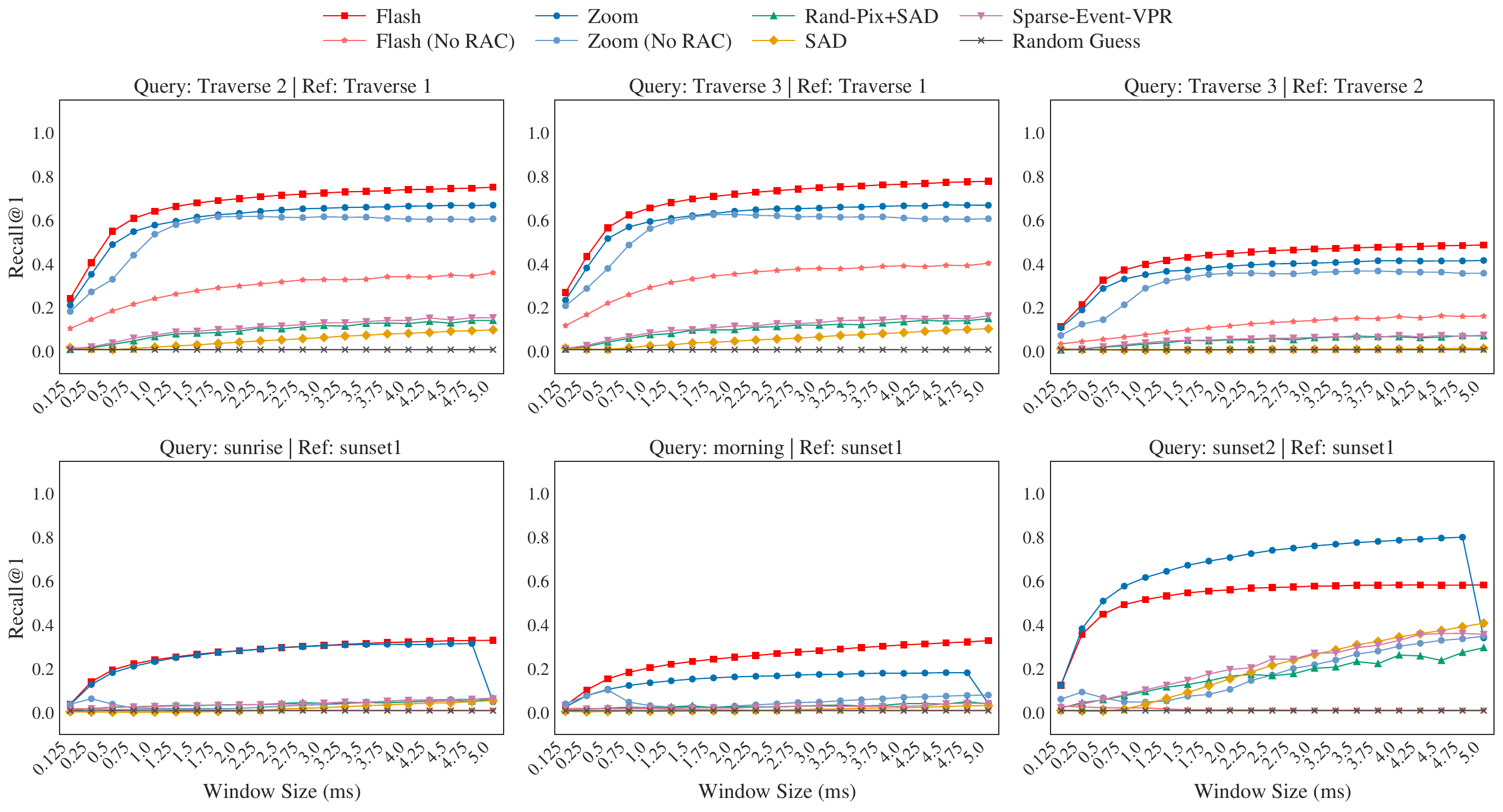}%
  \vspace*{-0.5cm}
  \caption{Recall@1 across different event accumulation window sizes for the QCR-Event-VPR dataset (top row) and the Brisbane-Event-VPR dataset (bottom row). Flash and Zoom consistently achieve the highest recall. At sub-millisecond timescales, Flash improves matching performance by 11.33× (QCR) and 5.92× (Brisbane), while Zoom outperforms the baseline by 10.02× (QCR) and 5.89× (Brisbane).}
  \vspace{-0.5cm}
  \label{fig:r1}
\end{figure*}

\subsection{Baselines}
\label{subsubsec:baselines}

We benchmark our method against existing event camera-based VPR approaches, which are not explicitly designed for sub-millisecond timescales. We also introduce a strong new baseline, Zoom, which leverages event count information along with active pixel locations for sub-millisecond VPR.

\subsubsection{Existing Methods}
\label{para:exist_methods}
In the following, we outline the set of baselines used for comparison before introducing our proposed variant. We first consider the sum of absolute differences (SAD) computed over all pixels in the frame, a widely used approach in prior works \cite{milford2015towards, 10802384}. Since we base our approach on sampling query pixels, we further evaluate two pixel sampling strategies: first, we sample a fixed set of random pixels across query-reference pairs. Distances are computed using SAD (Rand-Pix+SAD), following the baseline approach in \cite{9925670}. In the second, we use the variance-based pixel selection strategy proposed in \cite{9925670}, which selects pixels with high variance across the reference set and computes SAD over these locations (Sparse-Event-VPR). This approach, closely related to our proposed method, focuses on the most distinctive pixels while discarding less informative ones. Sparse-Event-VPR outperforms certain learning-based event VPR methods, including Event-VPR \cite{kong2022event} and EventVLAD \cite{lee2021eventvlad}. We sample 150 pixels for Rand-Pix+SAD and Sparse-Event-VPR as done in \cite{9925670}.

We note that reconstruction-based pipelines such as E2VID \cite{Rebecq19pami}, a key component of Ensemble-Event-VPR \cite{9201344}, cannot generate stable outputs at the fine temporal resolutions considered here and are therefore excluded from our quantitative evaluation.

\subsubsection{Zoom}
\label{para:zoom}

As a baseline closely aligned with the proposed method, we use event count frames instead of binary frames and compute a masked sum of absolute differences (SAD) over the query’s active pixels:
\begin{equation}
D(I_Q, I_R) = \sum_{(x,y) \in \Omega_{I_Q}} \lvert I_Q(x,y) - I_R(x,y)\rvert.
\end{equation}
We normalize the distances based on the relative activity of the query and reference frames: 
\begin{equation} w(I_Q, I_R) = \begin{cases} \dfrac{|\Omega_{I_R}|}{|\Omega_{I_Q}|} & |\Omega_{I_R}| \ge |\Omega_{I_Q}| \\ 1 & \text{otherwise} \end{cases}.
\label{eq:weight} 
\end{equation}

Using full event counts, Zoom provides a richer but less efficient representation, highlighting the Flash’s speed and computational efficiency, while serving as a baseline.

\section{Results}
\label{sec:results}

\begin{figure}[t]
  \centering
  \includegraphics[width=0.9\linewidth]{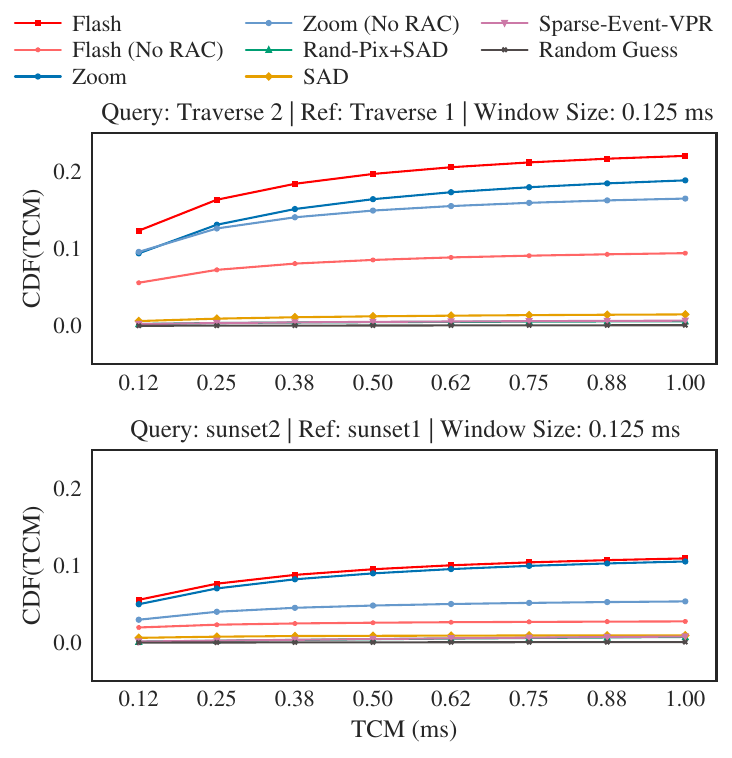}%
    \vspace*{-0.3cm}
  \caption{VPR performance measured using the Time between Correct Match (TCM) metric, where higher CDF values indicate more frequent correct recognitions. Flash and Zoom achieve remarkably higher correct matches within short intervals, with Flash exceeding the best baseline by up to 13.91$\times$ on QCR and 10.46$\times$ on Brisbane for a 0.125~ms event window.}
  \label{fig:ttc_sota}
  \vspace{-0.5cm}
\end{figure}

\subsection{Comparison to State-of-the-Art}
\label{subsec:sota}

We first evaluate the capabilities of existing event camera-based VPR approaches at sub-millisecond timescales, and then assess the performance of the proposed method against the baselines described in \cref{subsubsec:baselines}.

\paragraph{Recall@1}
\label{para:sota_r1}

\cref{fig:r1} presents the Recall@1 performance of Flash and Zoom compared to existing approaches across varying window sizes. Flash achieves the highest performance, improving average Recall@1 over all shown window sizes by $5.42\times$ i.e.~$\sim542$\% (QCR) and $2.67\times$ (Brisbane) over the best baseline (Sparse-Event-VPR), with even larger gains of $11.33\times$ (QCR) and $5.92\times$ (Brisbane) at sub-millisecond timescales. Zoom also delivers consistent improvements, with overall gains of $4.69\times$ (QCR) and $2.69\times$ (Brisbane). At sub-millisecond scales, Zoom achieves $10.02\times$ (QCR) and $5.89\times$ (Brisbane) gains. Between the two proposed methods, Flash outperforms Zoom on QCR by 0.13× overall and 0.12× at sub-millisecond scales. In contrast, Zoom overall slightly outperforms Flash on Brisbane ($0.01\times$) due to its stronger performance on the sunset1–sunset2 pair with no significant difference in performance at sub-millisecond timescales. These results indicate that the locations of active pixels provide more discriminative information than other pixel sampling strategies. Interestingly, Zoom’s performance drops at 5 ms on the Brisbane dataset due to RAC’s inability to fully mitigate aliasing from high-activity reference frames.

We also evaluate the importance of RAC in our proposed methods. On QCR, Flash and Zoom outperform their variants without RAC by $1.39\times$ and  $0.11\times$, respectively. On Brisbane which includes outdoor lighting challenges such as sun glare, the improvements of using RAC are larger: $23.87\times$ for Flash and $2.82\times$ for Zoom. At sub-millisecond scales, Flash improves by $1.89\times$ (QCR) and $9.25\times$ (Brisbane), while Zoom improves by $0.34\times$ (QCR) and $2.52\times$ (Brisbane). These results show that normalizing similarity scores based on relative query–reference activity is both beneficial and crucial under challenging illumination conditions.

\paragraph{Time to Correct Match (TCM)}
\label{para:sota_tcm}

Given our focus on high-speed localization, we evaluate position blind time using the TCM metric. \cref{fig:ttc_sota} shows that Flash and Zoom achieve correct position estimates far more frequently than the baselines. For VPR predictions with a 0.125 ms event window, Flash predicts the correct position on or before 1 ms $13.91\times$ more frequently on QCR and $10.46\times$ more frequently on Brisbane compared to the best baseline (SAD). Zoom also achieves substantial gains, outperforming the baseline by $11.77\times$ on QCR and $10.04\times$ on Brisbane. These results demonstrate that our proposed methods substantially reduce the time instant for which the robot operates without awareness of its position in the map.

\subsection{Temporal Lower Bound for Event Camera-based VPR}
\label{subsec:temp_lb}

\begin{figure}[t]
  \centering
  \includegraphics[width=0.9\linewidth]{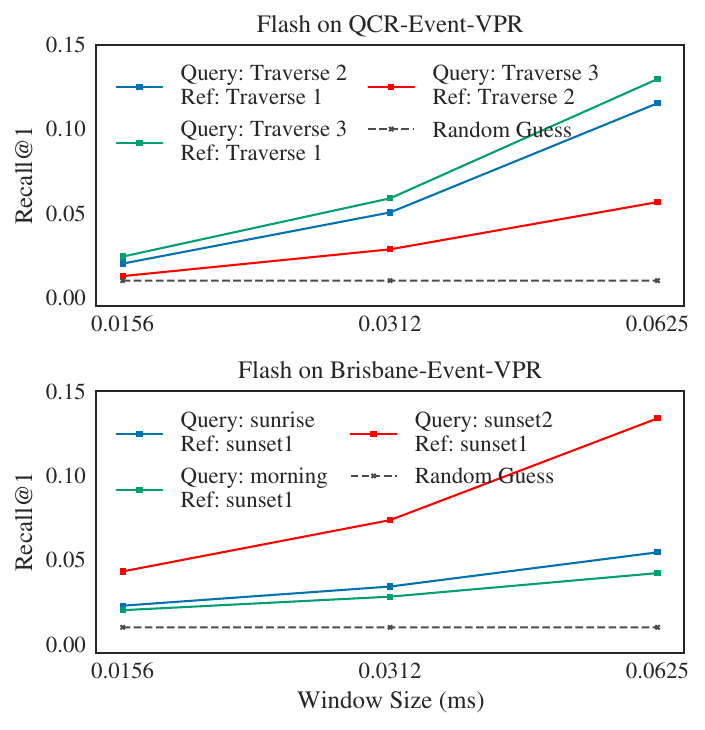}%
  \vspace*{-0.3cm}
  \caption{Recall@1 performance of Flash across ultra-short temporal accumulation windows. Even at $\sim$15~$\mu$s, corresponding to only a few tens of events and sub-millimetre robot motion, Flash achieves accuracy above random chance, demonstrating the feasibility of ultra-low latency event-based VPR and establishing a practical lower bound on the temporal resolution required.}
  \label{fig:ultra}
  \vspace{-0.5cm}
\end{figure}

To evaluate the minimal temporal requirements for event-based VPR, we plot Recall@1 as a function of accumulation window sizes on the order of tens of $\mu s$ (\cref{fig:ultra}). Even at the smallest window of $\sim$15~$\mu$s, our system achieves performance above random chance, demonstrating that ultra-low-latency VPR is feasible, albeit with reduced Recall@1. At such timescales, only a few to tens of events are available for matching (see \cref{subsubsec:ev_stat}), and the robot has moved just a fraction of a millimeter in both QCR and Brisbane. However, due to the sensing dynamics of event cameras, we note that some scene information from earlier motion appears within this interval. These results establish a practical lower bound on the temporal resolution required for event-based VPR, highlighting the ability of our method to extract discriminative spatial features from extremely short temporal slices.

\subsection{Sub-sampling the Reference Database}
\label{subsec:sub_sample}

\begin{figure}[t]
  \centering
  \includegraphics[width=0.9\linewidth]{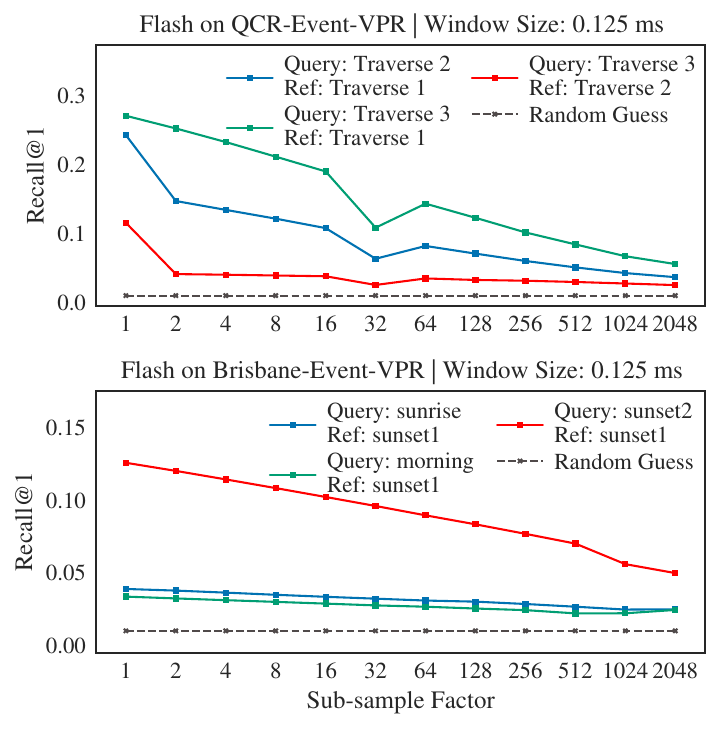}%
  \vspace*{-0.3cm}
  \caption{Effect of sub-sampling the reference database on Recall@1 for QCR and Brisbane. Progressive halving of the reference frames maintains Recall@1 with minimal loss while reducing storage and search time, with the first halving reducing Recall@1 by $\sim$36.85\% on QCR and $\sim$3.67\% on Brisbane. Aggressive reductions degrade performance, with total drops of $\sim$61.40\% and $\sim$23.56\% for a 2048× reduction.}
  \label{fig:subsampling}
\end{figure}

As our system operates at sub-millisecond latency, smaller temporal windows result in large reference databases. For 0.125 ms window size, this corresponds to $\sim$5.8M frames for Brisbane and $\sim$1.3M frames for QCR datasets. This leads to increased storage demands and longer search times, which are mitigated to a certain extent by our sparse and binary feature descriptors. To further reduce the compute, we progressively half the database size by retaining every second reference frame to tradeoff accuracy for lower latency. As shown in \cref{fig:subsampling}, the first halving reduces Recall@1 by $\sim$36.8\% on QCR and only $\sim$3.6\% on Brisbane, while cutting storage and search time by 50\%. Across successive halvings, the average Recall@1 drop per halving is $\sim$11.6\% for QCR and $\sim$4.8\% for Brisbane, resulting in total reductions of $\sim$61.4\% and $\sim$23.5\%, respectively, at a 2048× database reduction. These results highlight that moderate sub-sampling provides substantial efficiency gains with minimal accuracy loss.

\subsection{Event Data Statistics}
\label{subsubsec:ev_stat}

\begin{table}[t]
\caption{Average events and active pixels vs.~accumulation window size for QCR and Brisbane, showing that sub-ms VPR is possible from very sparse data.}
\label{tab:ev_stat}
\resizebox{\columnwidth}{!}{%
\begin{tabular}{ccccc}
\toprule
\multirow{2}{*}{\textbf{\begin{tabular}[c]{@{}c@{}}Window Size\\ (ms)\end{tabular}}} & \multicolumn{2}{c}{\textbf{QCR-Event-VPR}} & \multicolumn{2}{c}{\textbf{Brisbane-Event-VPR}} \\ 
\cmidrule(lr){2-3} \cmidrule(lr){4-5}
 & \textbf{\#Events} & \textbf{\#Active Pixels} & \textbf{\#Events} & \textbf{\#Active Pixels} \\ 
\midrule
0.0156 & 3.50 & 3.48 & 12.65 & 12.44 \\ 
0.0312 & 7.08 & 7.01 & 25.31 & 24.65 \\ 
0.0625 & 14.16 & 13.87 & 50.62 & 48.45 \\ 
0.1250 & 28.32 & 27.18 & 100.66 & 89.01 \\ 
0.2500 & 56.64 & 52.34 & 202.48 & 163.38 \\ 
0.5000 & 113.29 & 97.74 & 404.95 & 291.48 \\ 
0.7500 & 169.93 & 137.80 & 607.43 & 399.66 \\ 
1.0000 & 226.58 & 173.58 & 809.90 & 493.01 \\ 
\bottomrule
\end{tabular}
}
\vspace{-0.5cm}
\end{table}

The datasets used in our experiments were collected using the DAVIS346 event camera, which has a sensor resolution of 346×260 pixels. We downsample the event frames to 86×45 before using them for VPR. \cref{tab:ev_stat} reports the average number of events and active pixels for different temporal accumulation windows in the QCR and Brisbane datasets. At sub-millisecond windows, only a small fraction of pixels are active. At 1 ms, $\sim$173 pixels (3.1\%) for QCR and $\sim$492 pixels (8.8\%) for Brisbane are active on average per frame, whereas at the smallest window of $\sim 15 \mu s$, only $\sim$3 pixels (0.05\%) for QCR and $\sim$12 pixels (0.2\%) for Brisbane are active on average per frame, and these are sufficient for Flash to achieve above-chance performance. As the window size decreases, the number of events per active pixel tends toward 1. This supports our choice of binary frames for sub-millisecond VPR.

\section{Discussion and Conclusions}
\label{sec:discussions}

We presented \emph{Flash}, a novel event-based visual place recognition method that departs from dense event representations and instead operates directly on active pixel locations for efficiency. By processing sub-millisecond slices of event data with lightweight operations, Flash enables ultra-low-latency localization. Our experiments show that Flash outperforms baselines by a large margin, demonstrating that accurate place recognition can be achieved directly from event locations at unprecedented latencies, opening new directions for time-critical robotic localization.

Despite these promising results, our evaluation remains limited to datasets that do not fully capture the challenges of high-speed robotics. In particular, the performance of Flash under extreme motion dynamics and very high event rates is yet to be explored.

Future work includes deploying Flash on dedicated hardware to fully exploit its computational efficiency, integrating it into event-based visual-inertial odometry pipelines for high-speed SLAM, and evaluating it on high-speed datasets such as drone racing or agile manipulation. Such extensions would further establish Flash as a practical component of real-world event-based perception systems.

\bibliography{bibfiles/references}
\bibliographystyle{styles/IEEEtran}

\end{document}